\documentclass[journal]{IEEEtran}
\usepackage{booktabs}
\usepackage{makecell}
\usepackage{multirow}
\usepackage{float}
\usepackage{afterpage} 
\usepackage{stfloats}
\usepackage{amsmath,amsfonts}
\usepackage{algorithmic}
\usepackage{algorithm}
\usepackage{array}
\usepackage[caption=false,font=normalsize,labelfont=sf,textfont=sf]{subfig}
\usepackage{textcomp}
\usepackage{stfloats}
\usepackage{url}
\usepackage{verbatim}
\usepackage{graphicx}
\usepackage{cite}
\usepackage{placeins} 
\begin{document}

\title{Unsupervised Image Super-Resolution Reconstruction Based on Real-World Degradation Patterns}

\author{Yiyang Tie, Hong Zhu, Yunyun Luo, Jing Shi~\IEEEmembership{}
\thanks{This work was supported in part by the Key Lab.of Manufacturing Equipment of Shaanxi Province, grant number JXZZZB-2022-02 and Natural Science Basic Research Program of Shaanxi, grant number 2021JQ-487.  (\textit{Corresponding author: Hong Zhu})}}

\markboth{IEEE TRANSACTIONS ON CIRCUITS AND SYSTEMS FOR VIDEO TECHNOLOGY}%
{Shell \MakeLowercase{\textit{et al.}}: A Sample Article Using IEEEtran.cls for IEEE Journals}

\maketitle

\begin{abstract}
The training of real-world super-resolution reconstruction models heavily relies on datasets that reflect real-world degradation patterns. Extracting and modeling degradation patterns for super-resolution reconstruction using only real-world low-resolution (LR) images remains a challenging task. When synthesizing datasets to simulate real-world degradation, relying solely on degradation extraction methods fails to capture both blur and diverse noise characteristics across varying LR distributions, as well as more implicit degradations such as color gamut shifts. Conversely, domain translation alone cannot accurately approximate real-world blur characteristics due to the significant degradation domain gap between synthetic and real data. To address these challenges, we propose a novel TripleGAN framework comprising two strategically designed components: The FirstGAN primarily focuses on narrowing the domain gap in blur characteristics, while the SecondGAN performs domain-specific translation to approximate target-domain blur properties and learn additional degradation patterns. The ThirdGAN is trained on pseudo-real data generated by the FirstGAN and SecondGAN to reconstruct real-world LR images. Extensive experiments on the RealSR and DRealSR datasets demonstrate that our method exhibits clear advantages in quantitative metrics while maintaining sharp reconstructions without over-smoothing artifacts. The proposed framework effectively learns real-world degradation patterns from LR observations and synthesizes aligned datasets with corresponding degradation characteristics, thereby enabling the trained network to achieve superior performance in reconstructing high-quality SR images from real-world LR inputs.
\end{abstract}


\section{Introduction}
\IEEEPARstart{I}{mage} super-resolution (SR) is a fundamental task in computer vision and image processing. Among its various branches, Single Image Super-Resolution (SISR) refers to the process of reconstructing a high-resolution (HR) image from a single low-resolution (LR) image. Deep learning-based super-resolution models typically require paired LR-HR images for training. However, real-world LR images are captured directly under practical conditions, making it difficult to obtain corresponding HR references and degradation information. Consequently, performing super-resolution using only unpaired real-world LR images remains highly challenging.

Many existing methods \cite{zhang2021designing,wang2021real} employ sophisticated synthetic degradation models to generate paired datasets that simulate real-world degradation patterns, aiming to train networks capable of performing super-resolution reconstruction on real-world LR images. However, the degradation patterns modeled by these synthetic algorithms often fail to accurately reflect the true degradation characteristics present in the real-world LR domain. In real-world scenarios, LR images captured under different devices and shooting conditions are subject to a variety of unknown degradations, including blur, noise, and color gamut degradation. If the training dataset exhibits significant discrepancies from the real-world degradation patterns of LR images, the network trained on such data will struggle to achieve satisfactory super-resolution performance on LR images from the real world LR domain (i.e., real-world LR images). Some methods \cite{kohler2019toward,joze2020imagepairs} construct real-world SR datasets using hardware setups and alignment algorithms, but the HR-LR pairs remain imperfectly aligned and expensive to obtain. As a result, real-world super-resolution datasets remain scarce throughout the development of super-resolution research. In most cases, only LR images are accessible, with no paired HR counterparts available. Achieving super-resolution reconstruction under this constraint is both technically challenging and of significant research interest. 

An ideal real-world super-resolution reconstruction model should first explore and model the degradation patterns inherent in real-world images. In the absence of HR references and with only real-world LR images available, mainstream methods for Real-world Single Image Super-Resolution (RSISR) can be categorized into two major approaches. The first category focuses on learning degradation patterns solely from internal datasets, targeting the degradation characteristics within the real-world LR images themselves.  For example, ZSSR \cite{shocher2018zero} requires no pre-training; instead, it performs super-resolution reconstruction by exploiting internal features of a single LR image during the testing phase. Similarly, KernelGAN \cite{bell2019blind} estimates the blur kernel by mining the internal degradation patterns of a single image.

The second category consists of methods based on external datasets, the majority of which rely on explicitly defined degradation models with mathematical formulations. Representative approaches include RealESRGAN \cite{wang2021real}, BSRGAN \cite{zhang2021designing}, among others. These methods explicitly construct the image degradation pipeline through mathematical equations, synthesizing LR images by combining blur, additive noise, and lossy compression artifacts through first-order or second-order degradation. The synthesized LR images differ from those in data sets like DIV2K \cite{agustsson2017ntire}, as their more complex degradation better reflects real-world scenarios. RealSR \cite{ji2020real} attempts to simulate real-world degradation patterns by constructing a large degradation kernel pool; however, accurately modeling the complex degradation patterns inherent in real-world LR images remains highly challenging. Liang et al. \cite{liang2021mutual} argued that a single image may correspond to multiple blur kernels and thus proposed a multi-blur-kernel modeling approach for a single image. DASR(Degradation-Aware SR) \cite{wang2021unsupervised}extract discriminative representations to obtain accurate degradation information. Domain translation-based models, such as CinCGAN \cite{yuan2018unsupervised}, transform real-world domain LR images into bicubic-downsampled LR domain images, achieving certain improvements. SwinIR \cite{liang2021swinir} was the first to introduce the Swin Transformer \cite{liu2021swin} into the field of super-resolution, improving the recovery of fine image details. HAT \cite{chen2023activating} incorporated an Overlapping Cross-Attention module, enabling finer detail restoration under complex degradation scenarios. DiffIR \cite{xia2023diffir} introduces diffusion models into the image restoration field, significantly enhancing detail reconstruction. However, it adopts exactly the same degradation pipeline as Real-ESRGAN for data synthesis, which limits its applicability to real-world degradation scenarios.

Most existing real-world super-resolution models rely heavily on datasets that incorporate realistic degradation patterns for training. However, current data collection or synthesis methods struggle to adapt to LR images captured under varying imaging devices and acquisition conditions. As a result, the constructed datasets fail to accurately model the diverse blur characteristics, noise distributions, and more implicit degradation patterns—such as color gamut degradation—present in real-world scenarios.

A key challenge in real-world super-resolution is learning diverse degradation patterns under a large blur domain gap between synthetic and real LR images. Designing a generalized synthesis pipeline that produces degradation-reflective datasets tailored to real-world scenarios remains crucial for effective SR.     

Based on an analysis of the degradation patterns in real-world images, this paper proposes an unsupervised super-resolution reconstruction method. Since our approach employs three GAN networks, it is named TripleGAN. Learning the degradation characteristics of real-world LR images is the central objective of the entire network pipeline. FirstGAN introduces a cascaded blur kernel module based on the internal blur self-similarity of images, producing an intermediate domain dataset that initially approaches the blur degradation domain of real-world LR images. SecondGAN maps this intermediate domain to the real-world LR domain, further approximating the blur characteristics of real-world LR images while learning other degradation patterns such as noise degradation and color gamut degradation, thereby generating a pseudo-real domain dataset that reflects real-world degradation characteristics. ThirdGAN is used to train the pseudo-real domain dataset and to achieve super-resolution reconstruction of real-world LR images. This method enables unsupervised super-resolution reconstruction using only real-world LR images and HR images from the unrelated DIV2K dataset. Qualitative and quantitative analyses conducted on the RealSR \cite{cai2019toward} and DrealSR \cite{wei2020component} datasets demonstrate that our method achieves the best performance among the compared methods in the experiments.

The main contributions of our method are summarized as follows:
1) We analyze the degradation patterns of real-world images and identify that real-world degradation includes not only noise and blur but also color gamut degradation. We further verify that datasets accurately reflecting real-world degradation patterns play a dominant role in the super-resolution reconstruction of real-world images.
2) We propose a cascaded kernel module based on the internal blur self-similarity of images to generate a blur kernel pool for real-world LR images, thereby constructing a transition domain dataset that initially captures the blur degradation characteristics of real-world LR images.
3) We introduce a domain adaptation approach to map the transition domain dataset toward the real-world LR domain in the feature space, enabling more accurate learning of real-world LR blur characteristics, along with diverse degradation patterns such as noise, color gamut shifts, and other implicit degradations. This ultimately leads to the construction of a pseudo-real domain dataset that better reflects the degradation patterns of real-world LR images.

We observe that in real-world super-resolution reconstruction, the dataset plays a dominant role in determining reconstruction performance. A dataset that accurately reflects real-world degradation patterns can enable even ESRGAN-based models to achieve high-quality super-resolution results better than Transformer-based models or Diffusion-base models. Our TripleGAN framework has strong transferability, as the pseudo-real domain dataset generated by FirstGAN and SecondGAN can be used to train a wide range of super-resolution models. Moreover, under unsupervised evaluation metrics, our method outperforms various super-resolution models based on Diffusion and Transformer architectures, highlighting the critical role of accurately modeling real-world degradation patterns—especially in the absence of real-world HR images.

\section{Related Works}
\subsection{Single Image Super Resolution}
The task of SISR entered the deep learning era with the introduction of SRCNN \cite{dong2014learning} by Dong et al., which was the first to apply convolutional neural networks to the SISR problem. In 2016, Kim et al. \cite{kim2016accurate} proposed the Very Deep Super-Resolution Network (VDSR), which improved reconstruction quality by increasing network depth. In 2017, Tong et al. proposed Dense Skip Connections \cite{tong2017image} to further improve learning capacity through dense connections. Zhang et al. \cite{zhang2018image} incorporated channel attention mechanisms to enhance the local correlation of features.  

With the rise of Transformer architectures, researchers have explored the application of Transformers in super-resolution reconstruction. In 2019, Liang et al. introduced SwinIR \cite{liang2021swinir}, a Swin Transformer \cite{liu2021swin} based image restoration method that employed hierarchical feature extraction and window-based self-attention mechanisms, achieving notable improvements in reconstruction quality. Notably, the Dual Aggregation Transformer (DAT) \cite{chen2023dual} introduces a hybrid design that combines spatial and channel self-attention. Chen et al. proposed HAT \cite{chen2023activating}, which integrated channel attention and window-based self-attention to enhance restoration across broader regions. This approach effectively captured both global and local information in images, leading to higher-quality super-resolution results.

At the same time, with the rapid development of generative models, an increasing number of generative approaches have been applied to the super-resolution field to enhance the perceptual quality of reconstructed images. In 2017, SRGAN \cite{ledig2017photo} employed an adversarial training strategy, enabling the generator to produce high-resolution images with greater visual realism, while introducing perceptual loss \cite{johnson2016perceptual} to better optimize texture details. ESRGAN \cite{wang2018esrgan} improved the network architecture by introducing Residual-in-Residua Dense Blocks (RRDB) and a relativistic discriminator, further enhancing the sharpness and naturalness of the generated images. Wei et al. exploit large-scale GAN priors \cite{wei2024toward}to address extreme rescaling, leading to notable improvements in visual quality.

In recent years, diffusion models have achieved remarkable progress in the field of image super-resolution. Ho et al. introduced the Denoising Diffusion Probabilistic Models (DDPMs) \cite{ho2020denoising}, which established a solid theoretical foundation for diffusion models. Subsequently, Song et al. proposed a noise-conditional score network based on stochastic differential equations \cite{song2020score}. Moser et al. proposed the Diffusion Wavelet method \cite{moser2024waving}, which integrates discrete wavelet transforms with diffusion models. More recently, Gao et al. introduced an implicit diffusion model \cite{gao2023implicit} targeting continuous-scale super-resolution, enabling high-fidelity transformation from LR images to high-resolution outputs at arbitrary scales.

Current mainstream SISR models face significant challenges in real-world applications: their performance heavily relies on the consistency between the training data and the degradation characteristics of real LR images. If the training data fail to accurately reflect the complex degradation patterns—such as noise and blur—present in real images, the resulting reconstruction is prone to noticeable artifacts and distortions.

\subsection{Real-World Image Super Resolution}
There exists a significant domain gap between synthetic training data and real-world LR images. Models trained on such synthetic datasets often exhibit poor performance when applied to real-world scenarios. Therefore, constructing datasets that can effectively reflect the degradation patterns of real-world LR images has become a critical challenge in advancing super-resolution reconstruction. SRMD \cite{zhang2018learning} pioneered a degradation model with blur and noise beyond bicubic downsampling. MPFNet \cite{niu2024learning} adopts a similar strategy to better handle real-world degradations. To handle LR images synthesized with multiple blur kernels, IKC \cite{gu2019blind} was the first to adopt an iterative estimation strategy to better estimate the blur kernel. Building on this, Luo et al. proposed the DAN network \cite{huang2020unfolding}, which iteratively estimates degradation through a Restorer-Estimator loop. However, this approach assumes the availability of blur kernels during training, which are typically unknown in real-world LR images. BSRGAN\cite{zhang2021designing} proposed a degradation pipeline involving randomized combinations of blur, noise injection, downsampling, and JPEG compression. RealESRGAN \cite{wang2021real} further extended this by introducing higher-order degradation models and more complex blur and noise combinations. However, these methods rely on explicit and predefined degradation mechanisms during dataset synthesis, which still leave a notable domain gap from the unknown and implicit degradations found in real-world scenarios. KernelGAN, proposed by Bell-Kligler et al.\cite{bell2019blind}, tackled this challenge by learning internal degradation patterns directly from a single image to extract its blur kernel. RealSR \cite{ji2020real} extended this idea by introducing a noise pool on top of kernel estimation, achieving first place in the NTIRE 2020 challenge. Built upon BSRGAN and Real-ESRGAN, DASR (degradation-adaptive super-resolution) \cite{liang2022efficient} further introduces an efficient super-resolution framework specifically designed to handle varying levels of image degradation. Fan et al. proposed the ADASSR \cite{fan2024towards} module, which enhances generalization by integrating external degradation priors into the modeling of internal degradation patterns. Sun et al. introduced the SDFlow module \cite{sun2024learning}, which treats the degradation model and the super-resolution model as highly correlated components, and jointly trains them to strengthen their internal consistency. SUPIR \cite{yu2024scaling} utilizes the Real-ESRGAN degradation process on a large-scale corpus of tens of millions of HR images and incorporates textual annotations to enable multi-modal learning, thereby effectively training the super-resolution network with rich image-text semantic associations.

With the introduction of diffusion models, Yue et al. proposed ResShift \cite{yue2023resshift}, which has demonstrated promising performance in real-world image super-resolution tasks. The emergence of T2I-Adapter \cite{mou2024t2i} has significantly enhanced the controllability of text-to-image generation models by enabling effective conditioning on additional structural cues such as segmentation maps, edges, or depth, thereby improving generation quality and alignment with input prompts. When incorporated into super-resolution frameworks, this module provides stronger and more semantically aligned textual guidance, leading to improved generation fidelity and content consistency. Other Diffusion-based super-resolution models such as StableSR \cite{wang2024exploiting}, PASD \cite{yang2024pixel}, DiffBIR \cite{lin2024diffbir}, and SeeSR \cite{wu2024seesr} aim to effectively capture the semantic and textual information embedded in real-world degraded LR images. However, the LR images used for training in all these models are still synthesized using degradation pipelines based on Real-ESRGAN, which remains one of the most effective handcrafted degradation synthesis methods to date, rather than being derived from authentic real-world degradation processes.

To better capture the degradation patterns of real-world LR images, Cai et al.\cite{cai2019toward} focused on constructing real-world datasets. They used Canon 5D3 and Nikon D810 cameras to capture images of the same scene with different focal lengths using the same camera, thereby obtaining corresponding HR and LR image pairs. FSN \cite{guan2024frequency} directly trains on these real-world pairs. However, due to the challenges and cost in collecting HR images accurately aligned with real-world LR data, researchers shifted their attention toward methods that rely solely on real-world LR images. Yuan et al.\cite{yuan2018unsupervised} proposed CinCGAN, a “Cycle-in-Cycle” framework based on generative adversarial networks. Their method maps real-world LR images to the bicubic domain before applying a super-resolution model trained on bicubic degradation. Similarly, Chen introduced CycleSR \cite{chen2020unsupervised}, a two-stage framework with an indirect supervision pathway. ReDegNet \cite{li2022face} leverages the strong structural priors of real-world LQ (low-quality) face images and their restored HQ (high-quality) counterparts to learn degradation-aware, content-independent representations, which are then transferred to natural images for realistic degradation synthesis. Lee et al. proposed MSSR \cite{lee2022learning}, which introduces a probabilistic degradation generator based on a hierarchical latent variable model to learn the degradation distribution. SRTTA \cite{deng2023efficient} addresses the domain gap by introducing a test-time adaptation framework that dynamically adapts SR models to unknown degradations using a second-order degradation scheme guided by a degradation classifier. RealDGen \cite{peng2024towards} use a DDPM model to generated realistic-style dataset. However, relying solely on domain translation models is insufficient to bridge the significant gap between synthetic and real-world degradations, leaving a mismatch with the true characteristics of real-world LR images.

\section{Method}
\subsection{Real-World Image Degradation Analysis}
Acquiring paired HR and LR images with precise alignment in real-world scenarios is extremely challenging and costly. Under the current setting of real-world unsupervised super-resolution, where only unpaired real-world LR images are available, existing studies commonly utilize the HR images in DIV2K dataset \cite{agustsson2017ntire} to synthesize datasets that emulate the degradation patterns of real-world LR images. In unsupervised real-world image super-resolution research, we identify the core challenge as constructing a dataset that accurately reflects the actual real-world LR degradation process. Specifically, we argue that it is essential to jointly estimate key degradation factors, including blur kernels, color space shifts, and noise, which collectively define the complete degradation parameter space (i.e., degradation domain). When the constructed degradation domain model can closely approximate the real-world degradation characteristics in terms of blur, noise, and color shifts, a network trained under such degradation conditions will exhibit significantly enhanced adaptability to complex real-world scenarios, thereby achieving higher-quality reconstruction results.

We argue that conventional domain translation methods, which take bicubic downsampling LR images as input, fail to bridge the significant degradation domain gap between the real-world LR degradation domain and the bicubic degradation domain. In particular, the large disparity in blur degradation leads to limited effectiveness when using a single domain translation model, as the generated synthetic LR images exhibit blur characteristics inconsistent with real-world LR. Moreover, real-world LR images from different scenes exhibit diverse degradation patterns in terms of noise and color space shifts and large blur degradation domain gaps, making it extremely difficult to learn the degradation patterns of various real-world scenarios using a single degradation pattern extraction method, such as those that estimate blur kernels directly from the LR image or domain translation method. Consequently, neither a single degradation modeling approach nor a single domain translation model can effectively synthesize a dataset that accurately reflects real-world degradation. To address the challenge of significant degradation domain discrepancies, we propose a progressive degradation domain modeling strategy, which consists of two stages:

1) Primary Degradation Modeling: In the absence of paired high-resolution data, we propose a network that explicitly estimates the blur kernel by exploiting the internal self-similarity of blur characteristics within a single real-world LR image. The estimated blur kernel is then used to synthesize a dataset that preliminarily narrows the blur degradation gap toward the real-world LR images, thus constructing a Transitional Domain Dataset.

2) Degradation Domain Adaptation Enhancement: We introduce an adversarial domain adaptation network based on CycleGAN to map the intermediate dataset toward the real-world LR domain in the feature space. Through unsupervised learning under cycle-consistency constraints, and supported by the proposed Transitional Domain Dataset, the model further reduces the blur degradation discrepancy with the real-world LR domain while simultaneously learning the degradation patterns of real-world LR images, including noise characteristics and color space shifts. This process yields a Pseudo-real Domain Dataset aligned with the degradation distribution of real-world LR.

In Chapter 4, we analyze the degradation discrepancies between the Transitional Domain Dataset, the Pseudo-real Domain Dataset, and the real-world LR domain of real-world LR images.

\subsection{Primary Degradation Modeling}
Inspired by KernelGAN, we exploit the maximal inter-patch blur self-similarity within a single real-world LR image, leveraging FirstGAN in our proposed TripleGAN framework to estimate the ×2 blur kernel.

The blur kernel pool is constructed by cropping patches $a\sim {p_{data({I_{LR}})}}$ from the real-world low-resolution image ${I_{LR}}$. The objective of the generator ${G_1}$ is to generate a blur kernel ${k}$ such that, when ${I_{LR}}$ is downsampled and blurred by ${k}$, it becomes indistinguishable from $a\sim {p_{data({I_{LR}})}}$ in terms of blur characteristics. The discriminator ${D_1}$ maximizes the adversarial loss $L_{{\rm{ad}}v}^{GA{N_1}}$ of FirstGAN by increasing its score for real patches ${D_1}(a)$ while decreasing the score for generated patches $(a*{\rm{k}}){ \downarrow _2}$, there by forcing ${G_1}$ to improve the quality of ${k}$. The generator ${G_1}$ is minimized by adjusting ${k}$ so that the degree of fuzziness approximates the real block, thus enhancing the probability value of ${D_1}(((a*{\rm{k}}){ \downarrow _2}*k){ \downarrow _2})$. The adversarial loss formula for FirstGAN $L_{{\rm{ad}}v}^{GA{N_1}}$ is as follows:
\begin{equation}
\label{deqn_ex1}
\begin{split}
L_{{\rm{ad}}v}^{GA{N_1}} =\ 
& {E_{a\sim{p_{data({I_{LR}})}}}}[\log {D_1}(a)] \\
+\
& {E_{a\sim {p_{data({I_{LR}})}}}}[\log (1 - {D_1}((a*{\rm{k}}){ \downarrow _2}))]
\end{split}
\end{equation}

The blur kernel obtained from FirstGAN corresponds to a ×2 blur kernel, which is used to construct the transitional domain blur pool. To effectively obtain the ×4 LR image, we design a cascaded blur scheme. Specifically, the high-resolution image ${I_{HR}}$ is sequentially degraded by applying the ×2 blur kernel ${k_2}$ twice, followed by downsampling, to generate the ×4 low-resolution image ${I_{L{R_4}}}$. The process can be formulated as Equation (\ref{deqn_ex2}):
\begin{equation}
\label{deqn_ex2}
{I_{L{R_4}}} = {I_{HR}}*{k_4}{ \downarrow _4} = ({I_{HR}}*{k_2}{ \downarrow _2})*{k_2}{ \downarrow _2}
\end{equation}
where ${ \downarrow _2}$ and ${ \downarrow _4}$ denote downsampling operations by a factor of 2 and 4, respectively.

We synthesize a transitional Domain Dataset by combining 800 HR images from the DIV2K training set with the transition domain blur kernel pool constructed in our method. Although the resulting transition domain dataset already captures degradation characteristics of real-world LR images, a domain gap still exists between the blur degradation patterns of the transition domain and the real-world LR domain due to the inherent complexity of real-world degradation. In addition, the synthesized dataset lacks other degradation factors such as noise and color space shifts.

We synthesize a transitional Domain Dataset by combining 800 HR images from the DIV2K training set with the transition domain blur kernel pool constructed in our method. Although the resulting transition domain dataset already captures degradation characteristics of real-world LR images, a domain gap still exists between the blur degradation patterns of the transition domain and the real-world LR domain due to the inherent complexity of real-world degradation. In addition, the synthesized dataset lacks other degradation factors such as noise and color space shifts.

\subsection{Degradation Domain Adaptation Enhancement}
The degradation distribution of real-world LR images exhibits highly complex and ill-posed characteristics, with modeling challenges arising from the deep coupling of multiple degradation factors, making explicit modeling difficult \cite{luo2023end}. We argue that the Transitional Domain Dataset obtained through the initial degradation modeling already possesses certain characteristics of the real-world LR degradation patterns; however, there still exists a noticeable gap between it and the true degradation domain.

Next, we introduce a CycleGAN-based adversarial domain adaptation network to map the Transitional Domain Dataset towards the real-wrold LR domain in the feature space. We feed the Transitional Domain Dataset and the LR images from the RealSR dataset into CycleGAN as the training data, enabling the network to learn the characteristics of the real degradation domain and transform the transitional degradation domain toward the real-world domain, thereby generating the Pseudo-real Domain Dataset. The adversarial loss of the generator $L_{adv}^{GA{N_2}}$ is defined as:
\begin{equation}
\label{deqn_ex3}
\begin{split}
L_{adv}^{GA{N_2}} = \
&{{\rm E}_{I_{L{R_4}}^{real} \sim {p_{{\rm{data}}}}\left( {I_{L{R_4}}^{real}} \right)}}\left[ {{{\left( {{D_3}\left( {I_{L{R_4}}^{real}} \right) - 1} \right)}^2}} \right] \\ 
+\
&{{\rm E}_{I_{L{R_4}}^{trans} \sim {p_{{\rm{data}}}}\left( {I_{L{R_4}}^{trans}} \right)}}\left[ {{D_3}{{\left( {{G_2}\left( {I_{L{R_4}}^{trans}} \right)} \right)}^2}} \right]
\end{split}
\end{equation}
where $n$ denotes the transitional domain and   denotes the real-world LR domain. The generator ${G_2}$ learns to map images from the transitional domain to the real-world LR domain, while the discriminator ${D_3}$ evaluates the quality of the generated images to distinguish them from real-domain images.

The loss of cyclic consistency $L_{cyc}^{GA{N_2}}$ is:
\begin{equation}
\label{deqn_ex4}
\begin{split}
L_{cyc}^{GA{N_2}} =\
& {{\rm E}_{I_{L{R_4}}^{real} \sim {p_{{\rm{data}}}}\left( {I_{L{R_4}}^{real}} \right)}}\left[ {\parallel {G_2}\left( {{G_3}\left( {I_{L{R_4}}^{real}} \right)} \right) - I_{L{R_4}}^{real}{\parallel _1}} \right] \\
+\
& {{\rm E}_{I_{L{R_4}}^{trans} \sim {p_{{\rm{data}}}}\left( {I_{L{R_4}}^{trans}} \right)}}\left[ {\parallel {G_3}\left( {{G_2}\left( {I_{L{R_4}}^{trans}} \right)} \right) - I_{L{R_4}}^{trans}{\parallel _1}} \right]
\end{split}
\end{equation}
Where the generator ${G_3}$ learns to shift from real-world LR domain images to transitional domain images.

To ensure the transformation of the color gamut, we also introduce the identity loss $L_{identity}^{GA{N_2}}$ as suggested in CycleGAN, formulated as:
\begin{equation}
\label{deqn_ex5}
\begin{split}
L_{identity}^{GA{N_2}} =\
&{{\rm E}_{I_{L{R_4}}^{real} \sim {p_{{\rm{data}}}}\left( {I_{L{R_4}}^{real}} \right)}}\left[ {\parallel {G_2}\left( {I_{L{R_4}}^{real}} \right) - I_{L{R_4}}^{real}{\parallel _1}} \right] \\
+\
&{{\rm E}_{I_{L{R_4}}^{trans} \sim {p_{{\rm{data}}}}\left( {I_{L{R_4}}^{trans}} \right)}}\left[ {\parallel {G_3}\left( {I_{L{R_4}}^{trans}} \right) - I_{L{R_4}}^{trans}{\parallel _1}} \right]
\end{split}
\end{equation}

In summary, the final loss $L_{total}^{GA{N_2}}$  of the degradation domain adaptation enhancement model is the weighted sum of these four loss terms, formulated as:
\begin{equation}
\label{deqn_ex6}
\begin{split}
L_{\text{total}}^{GA{N_2}} =\
& L_{\text{adv}}^{GA{N_2}}\left( G_2 \right) + L_{\text{adv}}^{GA{N_2}}\left( G_3 \right) \\
+\
& \lambda_{\text{cyc}}^{GA{N_2}} \cdot L_{\text{cyc}}^{GA{N_2}}\left( G_2, G_3 \right) \\
+\
& \lambda_{\text{identity}}^{GA{N_2}} \cdot L_{\text{identity}}^{GA{N_2}}\left( G_2, G_3 \right)
\end{split}
\end{equation}
where $\lambda _{cyc}^{GA{N_2}}$ and $\lambda _{identity}^{GA{N_2}}$ denote the cycle-consistency loss coefficient and the identity mapping loss coefficient, respectively.

This approach not only enables a more accurate learning of the blur degradation pattern through FirstGAN, but also facilitates the learning of the noise degradation characteristics in real-world LR images, as well as more implicit degradation patterns such as color space degradation.

\subsection{Network Architecture}
\begin{figure*}[!t]
\centering
\includegraphics[width=\textwidth]{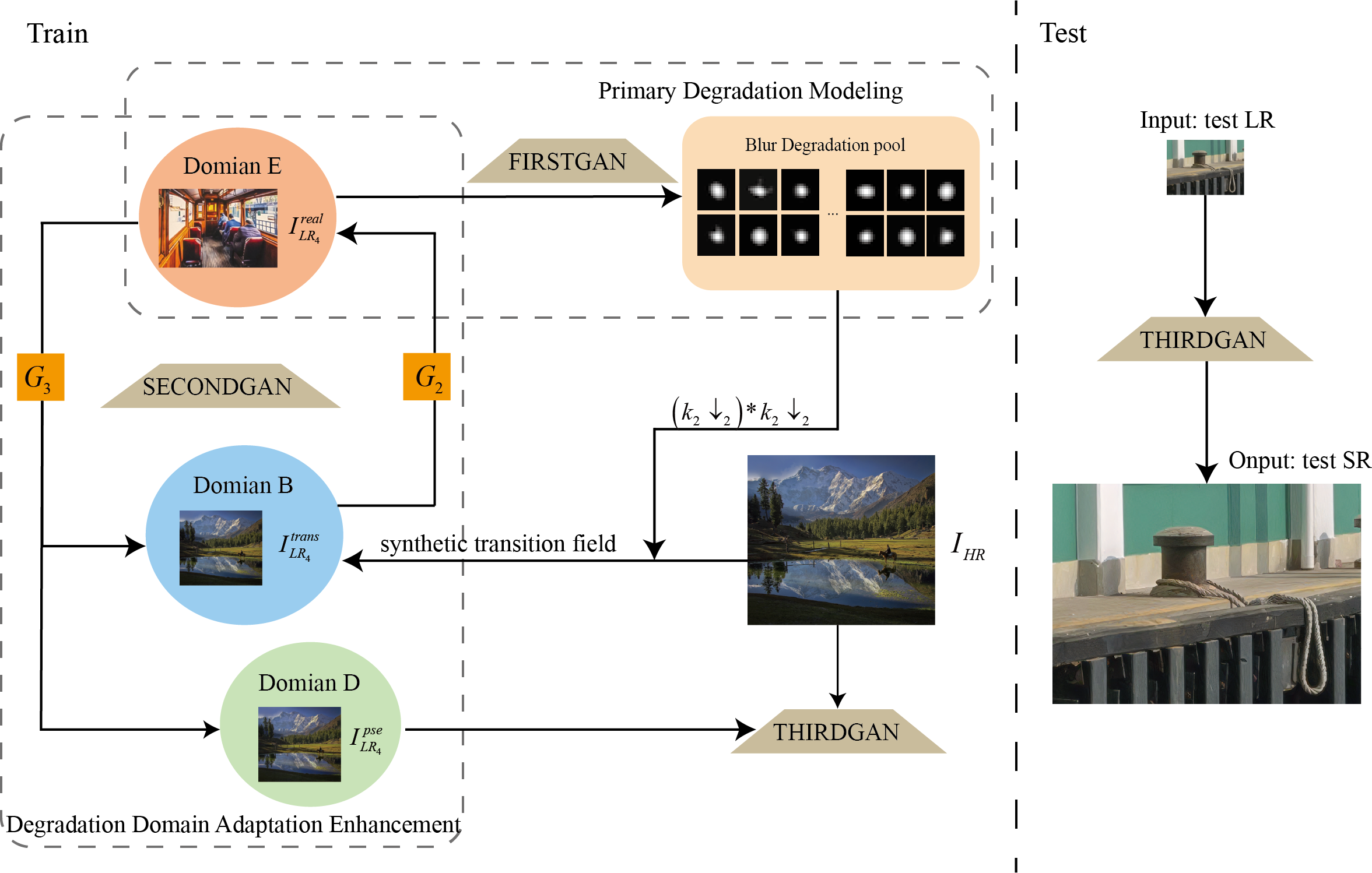}
\caption{The architecture of the proposed TripleGAN framework.}
\label{fig_1}
\end{figure*}

The TripleGAN model is composed of three GAN networks, with the overall structure illustrated in figure \ref{fig_1}. The FirstGAN constructs the Transitional Degradation Domain Dataset, which serves as the basis for the SecondGAN to perform domain translation, ultimately yielding the Pseudo-real Degradation Domain Dataset. FirstGAN performs primary degradation modeling by extracting blur kernels from real-world LR images to construct a blur kernel pool. This pool is then applied to high-quality images from the DIV2K dataset to synthesize a degradation domain that initially approximates the blur characteristics observed in real-world degradation.

SecondGAN achieves degradation domain adaptation and enhancement, mapping from the Transitional Degradation Domain Dataset to the real-world LR degradation domain. The ThirdGAN module adopts the ESRGAN architecture \cite{wang2018esrgan} and is trained on the Pseudo-Real Degradation Domain Dataset, enabling it to perform super-resolution on real-world LR images during inference. As illustrated in figure \ref{fig_1}, FirstGAN and SecondGAN collaboratively construct the pseudo-real domain and construct a dataset that closely aligns with the real-world degradation domain. This transition-domain-based domain transfer significantly reduces the domain gap compared to directly performing domain transfer, leading to a pseudo-real domain dataset whose LR images better approximate the degradation patterns of the real-world LR domain.

\section{Experiment}
Considering the difficulty of acquiring paired HR images in real-world scenarios for both training and evaluation, our method relies solely on ×4 downsampled LR images from real-world datasets. The evaluation is conducted entirely using unsupervised metrics, enabling practical and generalizable evaluation of real-world SR models. We first introduce the real-world datasets and evaluation metrics. In the experiments, we perform ablation studies on FirstGAN and SecondGAN in the TripleGAN framework by training ThirdGAN separately on the datasets synthesized by each submodule. This enables a comprehensive evaluation of the individual contributions of FirstGAN and SecondGAN. Additionally, we conduct comparative experiments under the complete TripleGAN framework against representative methods including RealSR \cite{ji2020real}, RealESRGAN \cite{wang2021real}, SwinIR-Real \cite{liang2021swinir}, HAT-Real \cite{chen2023activating}, RealDAN-GAN \cite{luo2023end}, DiffIR-Real \cite{xia2023diffir}, AdaSSR \cite{fan2024towards} and other representative real-world super-resolution methods.

\subsection{Training Datasets and Evaluation Metrics}
We conduct training on the ×4 downsampled datasets of RealSR Canon and RealSR Nikon, and perform testing on the ×4 downsampled datasets of RealSR Canon, RealSR Nikon, and DrealSR. Throughout the entire training process, we do not use any HR image information from these three datasets; only the LR images in the training datasets are used. The HR images from DIV2K used in our synthesized datasets are completely unrelated to the real-world LR images in the real-world LR domain—they are solely used to synthesize aligned image pairs for building the dataset.

Considering the unique nature of the real-world super-resolution reconstruction scenario—where it is often extremely difficult to obtain aligned HR images or reference HR images corresponding to the LR images in real-world settings—all evaluation metrics are based on no-reference (unsupervised) metrics to ensure better applicability to real-world use cases. To guarantee objectivity under no-reference evaluation, we adopt multiple unsupervised quality metrics, including NIQE \cite{xu2020unified}, PIQE \cite{mittal2012making}, BRISQUE \cite{venkatanath2015blind}, NRQM \cite{mittal2012no}, and PI \cite{ma2017learning}.

\subsection{Ablation Study}
We define five domains based on different generation methods to evaluate the effectiveness of each stage: \textbf{Domain A} (Standard Domain): Constructed using traditional bicubic interpolation; \textbf{Domain B} (Transitional Domain): Synthesized by applying blur kernels—extracted from real-world LR images via FirstGAN—to DIV2K HR images, forming a domain that partially reflects real-world blur degradation; \textbf{Domain C} (Independent Domain): Obtained by applying domain translation through SecondGAN on Domain A, shifting its distribution towards the real-world degradation domain; \textbf{Domain D} (Pseudo-real Domain):Built upon the degradation domain generated by FirstGAN, and further refined via domain translation using SecondGAN to progressively align with the real-world LR degradation distribution. This hierarchical synthesis ensures a closer match to real-world degradation patterns; \textbf{Domain E} (Real world-LR Domain): Composed of real-world LR images, serving as the evaluation reference.

\subsubsection{Module-wise Effectiveness Evaluation}
To verify the effectiveness of FirstGAN and SecondGAN in the TripleGAN framework, we conducted ablation experiments by setting up three experimental groups: FirstGAN-only: degradation pattern extraction using only FirstGAN; SecondGAN-only: Degradation domain transformation using only SecondGAN; Full TripleGAN: The complete framework (joint FirstGAN and SecondGAN). In all configurations, ThirdGAN is used for training and testing. The ablation results are shown in Table~\ref{tab1}, Table~\ref{tab2}.

Since all our weights were obtained by training on the RealSR dataset, the weights used for the DrealSR dataset are consistent with those obtained on the RealSR Nikon dataset.

\begin{table*}[!h]
\centering
\caption{Ablation study on the\textbf{ RealSR} ×4 test set \\
(best indicator is \textbf{bolded}, ↓ means lower is better, ↑ means higher is better).}
\label{tab1}
\begin{tabular}{lccccc|ccccc}
\toprule
\multirow{2}{*}{Method} & \multicolumn{5}{c|}{\textbf{Canon x4 test}} & \multicolumn{5}{c}{\textbf{Nikon x4 test}} \\
 & NIQE↓ & PIQE↓ & BRISQUE↓ & NRQM↑ & PI↓ & NIQE↓ & PIQE↓ & BRISQUE↓ & NRQM↑ & PI↓ \\
\midrule
FirstGAN+ThirdGAN   & 4.45 & 36.36 & 34.53 & 4.90 & 4.94 & 4.08 & 31.29 & 31.77 & 6.08 & 4.10 \\
SecondGAN+ThirdGAN  & 3.23 & 29.01 & 18.62 & 6.08 & 3.66 & 4.59 & 35.77 & 35.49 & 5.42 & 4.70 \\
TripleGAN           & \textbf{3.19} & \textbf{26.14} & \textbf{9.08}  & \textbf{6.92} & \textbf{3.15} & \textbf{3.63} & \textbf{29.19} & \textbf{10.46} & \textbf{6.89} & \textbf{3.41} \\
\bottomrule
\end{tabular}
\end{table*}

\begin{table*}[!h]
\centering
\caption{Ablation study on the \textbf{DRealSR} ×4 test set \\
(best indicator is \textbf{bolded}, ↓ means lower is better, ↑ means higher is better).}
\label{tab2}
\begin{tabular}{lccccc}
\toprule
Method & NIQE↓ & PIQE↓ & BRISQUE↓ & NRQM↑ & PI↓ \\
\midrule
FirstGAN+ThirdGAN   & 4.61 & 27.58 & 24.39 & 5.11 & 4.93 \\
SecondGAN+ThirdGAN  & 5.79 & 31.07 & 35.96 & 4.58 & 5.78 \\
TripleGAN           & \textbf{3.69} & \textbf{24.49} & \textbf{9.39} & \textbf{6.16} & \textbf{3.91} \\
\bottomrule
\end{tabular}
\end{table*}

We observed that using either FirstGAN or SecondGAN alone underperformed the combined framework, consistent with our hypothesis. Relying on a single model (for degradation pattern extraction or domain adaptation) exhibits limited efficacy in reducing the degradation domain gap and fails to approach the real-world LR domain. In contrast, the joint use of FirstGAN and SecondGAN improved all evaluation metrics. This stems from their complementary roles: FirstGAN primarily reduces the blur degradation domain gap, while SecondGAN further bridges the gap to the real-world LR blur domain and learns auxiliary degradation patterns (e.g., noise, color space distortion).

\subsubsection{Analysis of Blur Degradation Severity}
To evaluate the blur distance across different domains, we employed the Fourier high-frequency ratio for measurement. Firstly, the LR images from each domain were converted into grayscale images ${I_{{\rm{gray}}}}(x,y)$, where the grayscale image has a width $M$, height $N$, and $(x,y)$ is image pixel. A Fourier transform was then applied to ${I_{{\rm{gray}}}}(x,y)$ yielding $F(u,v)$, where $u$ and $v$ denote the frequency domain coordinates.
\begin{equation}
\label{deqn_ex7}
F(u,v) = \frac{1}{{MN}}\sum\limits_{x = 0}^{M - 1} {\sum\limits_{y = 0}^{N - 1} {{I_{{\rm{gray}}}}} } (x,y){e^{ - j2\pi \left( {\frac{{ux}}{M} + \frac{{vy}}{N}} \right)}}
\end{equation}

To facilitate the separation of low-frequency and high-frequency components, the frequency spectrum was first centered, resulting in a shifted frequency domain matrix ${F_{{\rm{shift}}}}(u,v)$.
\begin{equation}
\label{deqn_ex8}
{F_{{\rm{shift}}}}(u,v) = F\left( {u - \left\lfloor {\frac{M}{2}} \right\rfloor ,\;v - \left\lfloor {\frac{N}{2}} \right\rfloor } \right)
\end{equation}

Subsequently, the magnitude spectrum was extracted from the centered frequency spectrum.
\begin{equation}
\label{deqn_ex9}
|{F_{{\rm{shift}}}}(u,v)| = \sqrt {{\mathop{\rm Re}\nolimits} {{({F_{{\rm{shift}}}}(u,v))}^2} + {\mathop{\rm Im}\nolimits} {{({F_{{\rm{shift}}}}(u,v))}^2}} 
\end{equation}

Based on the magnitude spectrum, frequency band energy partitioning was performed. Specifically, the low-frequency mask, total energy ${E_{{\rm{total}}}}$, low-frequency energy ${E_{{\rm{low}}}}$, and high-frequency energy ${E_{{\rm{high}}}}$ were calculated accordingly,
\begin{equation}
\label{deqn_ex10}
Mask(x,y) = \left\{ {\begin{array}{*{20}{l}}
{1,}&{{\rm{if }}{{(x - {c_x})}^2} + {{(y - {c_y})}^2} \le {r^2}}\\
{0,}&{{\rm{otherwise}}}
\end{array}} \right.
\end{equation}

\begin{equation}
\label{deqn_ex11}
{E_{{\rm{total}}}} = \sum\limits_{u = 0}^{M - 1} {\sum\limits_{v = 0}^{N - 1} | } {F_{{\rm{shift}}}}(u,v)|
\end{equation}

\begin{equation}
\label{deqn_ex12}
{E_{{\rm{low}}}} = \sum\limits_{(u,v) \in Mask} | {F_{{\rm{shift}}}}(u,v)|
\end{equation}

\begin{equation}
\label{deqn_ex13}
{E_{{\rm{high}}}} = {E_{{\rm{total}}}} - {E_{{\rm{low}}}}
\end{equation}
where ${c_x} = N/2$, ${c_y} = M/2$.

Finally, the blur degree of the image was quantitatively measured by computing the high-frequency energy ratio SharpnessScore ${S_s}$.
\begin{equation}
\label{deqn_ex14}
{\rm{S_s}} = \frac{{{E_{{\rm{high}}}}}}{{{E_{{\rm{total}}}}}}
\end{equation}

We conducted an analysis on the ×4 super-resolution datasets of the RealSR dataset. Specifically, we used violin plots to visualize and compare the Fourier high-frequency ratio of LR images across Domain A, Domain B, Domain C, and Domain D, relative to the real-world LR domain (Domain E). This analysis aimed to evaluate the degree of similarity in blur degradation between the synthesized datasets under different configurations and the real-world LR domain.

\begin{figure}[!h]
\centering
\includegraphics[width=3.5in]{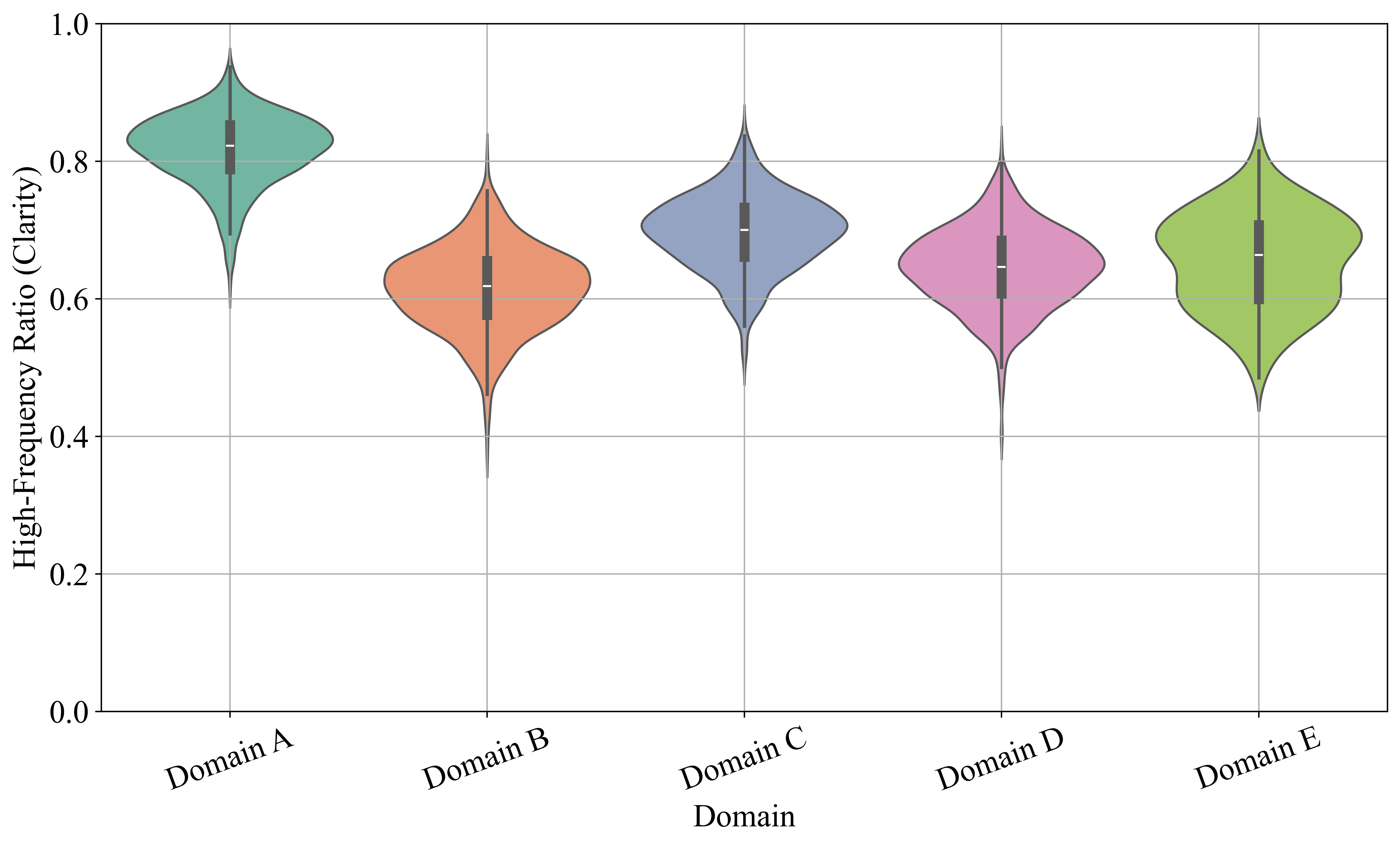}
\caption{High-frequency ratio distributions across different domains on the RealSR Canon dataset.}
\label{fig_2}
\end{figure}

\begin{figure}[!h]
\centering
\includegraphics[width=3.5in]{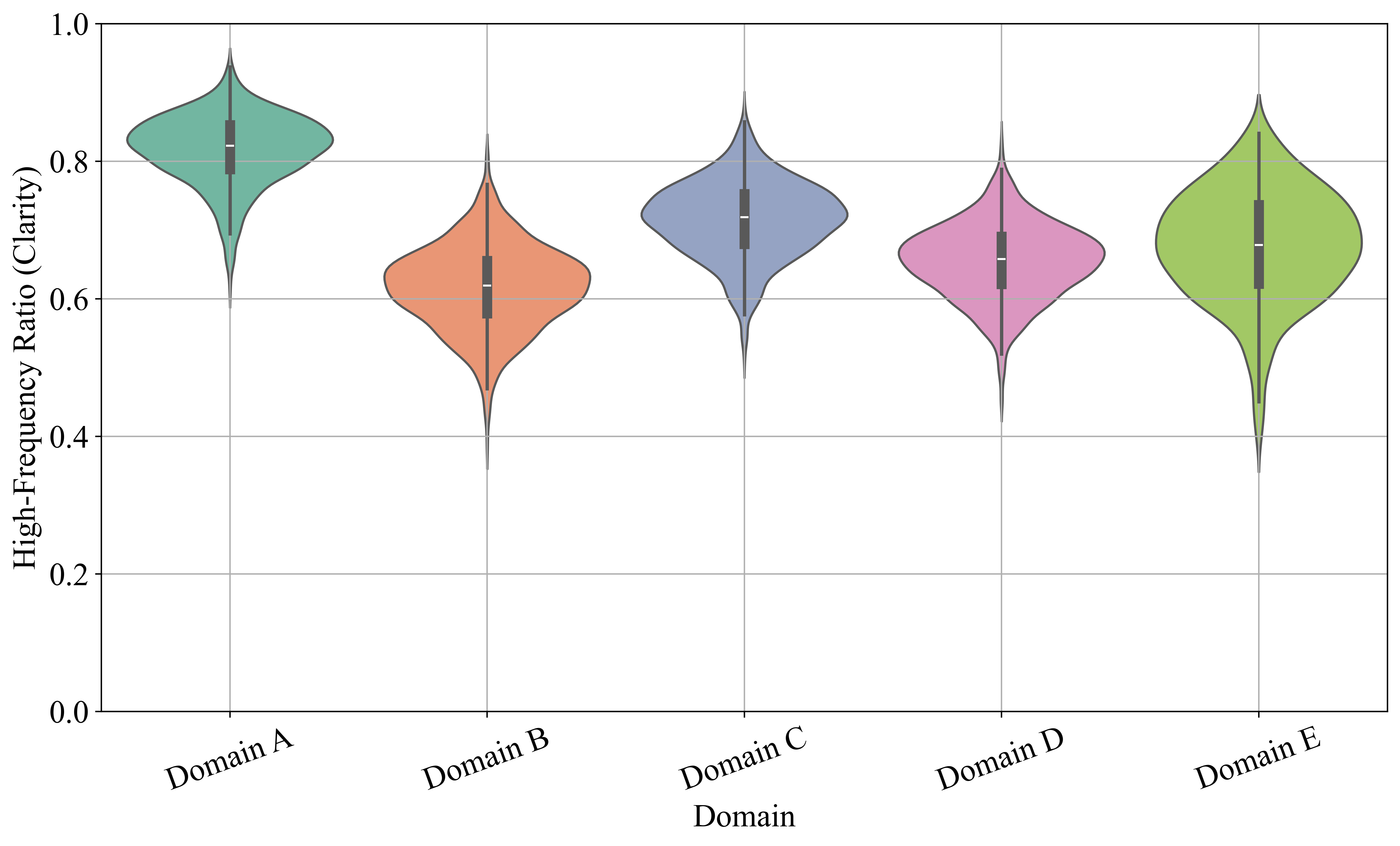}
\caption{High-frequency ratio distributions across different domains on the RealSR Nikon dataset.}
\label{fig_3}
\end{figure}

Experiments were conducted on the RealSR dataset, including the Canon and Nikon subsets, and the corresponding results are presented in figure \ref{fig_2} and figure \ref{fig_3}. In the resulting violin plots, the width of the violin represents the density distribution of the high-frequency ratio within each domain, i.e., the number of images exhibiting specific high-frequency ratio values. The gray bar inside each violin plot indicates the interquartile range, covering the middle 50\% of high-frequency ratio values within the domain. The white line within the gray bar marks the median value of the distribution.


Due to the inherent limitations of domain translation models, translating Domain A (bicubic interpolation domain) towards the real-world LR domain often results in suboptimal outputs, represented by Domain C. The significant gap in blur degradation characteristics between the bicubic and real-world domains hampers accurate translation. In contrast, when domain translation is applied to the transitional domain dataset (Domain B, synthesized by FirstGAN), the results exhibit blur characteristics that more closely resemble those of the real-world LR domain (Domain E).

Since Domain B is generated by convolving FirstGAN-extracted real-world blur kernels with HR images from the DIV2K dataset, it inherently lacks the noise patterns and color gamut properties of the real-world LR domain. We further analyze the noise distributions and color gamut shifts between Domain B and Domain D, where Domain D is synthesized by applying the SecondGAN-based domain translator to Domain B.

\subsubsection{Analysis of Noise Degradation Characteristics}
For the noise component, we first apply a Gaussian filter $Gus$ with a kernel size of 5 to the grayscale LR image ${I_{{\rm{gray}}}}(x,y)$, resulting in a smoothed image ${I_{smooth}}$,
\begin{equation}
\label{deqn_ex15}
{I_{{\rm{smooth}}}}\left( {x,y} \right) = \mathop \sum \limits_{i =  - k}^k \mathop \sum \limits_{j =  - k}^k I\left( {x,y} \right) \cdot Gus
\end{equation}

The noise estimate $Noise\left( {x,y} \right)$ is then calculated as the residual between the original image and the smoothed image ${I_{{\rm{smooth}}}}\left( {x,y} \right)$, in order to suppress structural content and reduce the interference of high-frequency details in noise estimation. The global standard deviation ${\sigma _{{\rm{noise}}}}$  of the residual serves as the estimated noise standard deviation, with ${\mu _{{\rm{noise}}}}$ denoting the residual mean and $W$, $H$ representing the width and height of the image, respectively, where $domain \in \{ B,D,E\}$.
\begin{equation}
\label{deqn_ex16}
Noise\left( {x,y} \right) = I\left( {x,y} \right) - {I_{{\rm{smooth}}}}\left( {x,y} \right)
\end{equation}

\begin{equation}
\label{deqn_ex17}
\begin{aligned}
\sigma_{domain}^{(i)} =\left( 
  \sqrt{ \frac{1}{HW} \sum\limits_{x,y} 
  \left( {\rm{Noise}}(x,y) - \mu_{{\rm{noise}}} \right)^2 }
\right)_{domain}^{(i)}
\end{aligned}
\end{equation}

\begin{equation}
\label{deqn_ex18}
{{\cal N}_{domain}} = \left\{ {\sigma _{domain}^{(1)},\sigma _{domain}^{(2)},...,\sigma _{domain}^{(i)}} \right\}
\end{equation}
Let $\sigma _{domain}^{\left( i \right)}$  represent the noise standard deviation of the $i$-th image within a given domain, and let the domain be one of Domain B, D, or E. Denote ${{\cal N}_{domain}}$ as the set of noise standard deviations in Domain B, Domain D, and Domain E.

We then compute the Wasserstein distance to compare the noise distribution differences between Domain B and Domain E - $W\left( {{{\cal N}_B},{{\cal N}_E}} \right)$ and between Domain D and Domain E - $W\left( {{{\cal N}_D},{{\cal N}_E}} \right)$, to verify whether the SecondGAN effectively learns the implicit noise degradation patterns of the real-world LR domain. The results of the noise distance analysis are summarized in Table~\ref{tab3}, $W\left( {{{\cal N}_B},{{\cal N}_E}} \right) = 2.41$, $W\left( {{{\cal N}_D},{{\cal N}_E}} \right) = 1.56$. Additionally, we perform a significance test on $W\left( {{{\cal N}_B},{{\cal N}_E}} \right) < W\left( {{{\cal N}_D},{{\cal N}_E}} \right)$ following the method in \cite{liang2021mutual}, with the resulting p-value approaching zero, strongly indicating that the improvement is statistically significant and validating the effectiveness of SecondGAN in learning real-world noise degradation.

\begin{table}[!h]
\centering
\caption{RealSR Dataset Noise Distance Analysis.}
\label{tab3}
\begin{tabular}{lccccc}
\toprule
Dataset & 
\makecell[c]{Noise Distance \\ $W\left( \mathcal{N}_B, \mathcal{N}_E \right)$} & 
\makecell[c]{Noise Distance \\ $W\left( \mathcal{N}_D, \mathcal{N}_E \right)$} & 
\makecell[c]{Significance Test \\ $p$} \\
\midrule
Canon & 2.41 & 1.56 & $p \approx 0.0000$ \\
Nikon & 2.58 & 1.54 & $p \approx 0.0000$ \\
\bottomrule
\end{tabular}
\end{table}

As shown, after domain transformation via the SecondGAN, the noise characteristics of Domain D become significantly closer to those of the real-world LR domain E, effectively narrowing the noise distribution gap.

\subsubsection{Analysis of Color Gamut Degradation}
We statistically analyzed the R, G, and B ratios of all pixels in Domains B, D, and E to verify whether the domain translation network the SecondGAN can learn implicit color gamut degradation. The color gamut distributions are presented in Table~\ref{tab4}.

\begin{table}[!h]
\centering
\caption{RealSR Dataset Color gamut distribution analysis \\ Channel Composition (R: G: B) in domain.}
\label{tab4}
\setlength{\tabcolsep}{4pt} 
\renewcommand{\arraystretch}{1.1} 
\begin{tabular}{lccccc}
\toprule
Dataset & Domain B & Domain D & Domain E \\
\midrule
Canon & 0.254: 0.338: 0.309 & 0.359: 0.341: 0.300 & 0.366: 0.340: 0.294 \\
Nikon & 0.353: 0.338: 0.309 & 0.360: 0.339: 0.301 & 0.368: 0.334: 0.299 \\
\bottomrule
\end{tabular}
\end{table}

As shown in the noise distance analysis in Table~\ref{tab5}, after transformation by the SecondGAN, the color distribution characteristics of Domain D become closer to the real-world LR domain E in most cases—except for the G channel on the RealSR Nikon dataset—indicating that the overall color gamut degradation shifts toward the real-world LR domain.

In summary, the combined framework of FirstGAN and SecondGAN in our proposed TripleGAN effectively models the degradation patterns of real-world LR images. FirstGAN reduces the blur domain gap, enabling SecondGAN to precisely capture the real-world LR domain’s blur characteristics. Meanwhile, SecondGAN learns the implicit degradation characteristics such as noise and color distribution. The pseudo-real domain dataset synthesized by FirstGAN and SecondGAN closely aligns with real-world LR degradation patterns. Based on this dataset, ThirdGAN—an enhanced variant of ESRGAN—achieves robust super-resolution reconstruction for real-world LR images.

\subsection{Comparative Experiments}
\begin{figure*}[t]
\centering
\includegraphics[width=\textwidth]{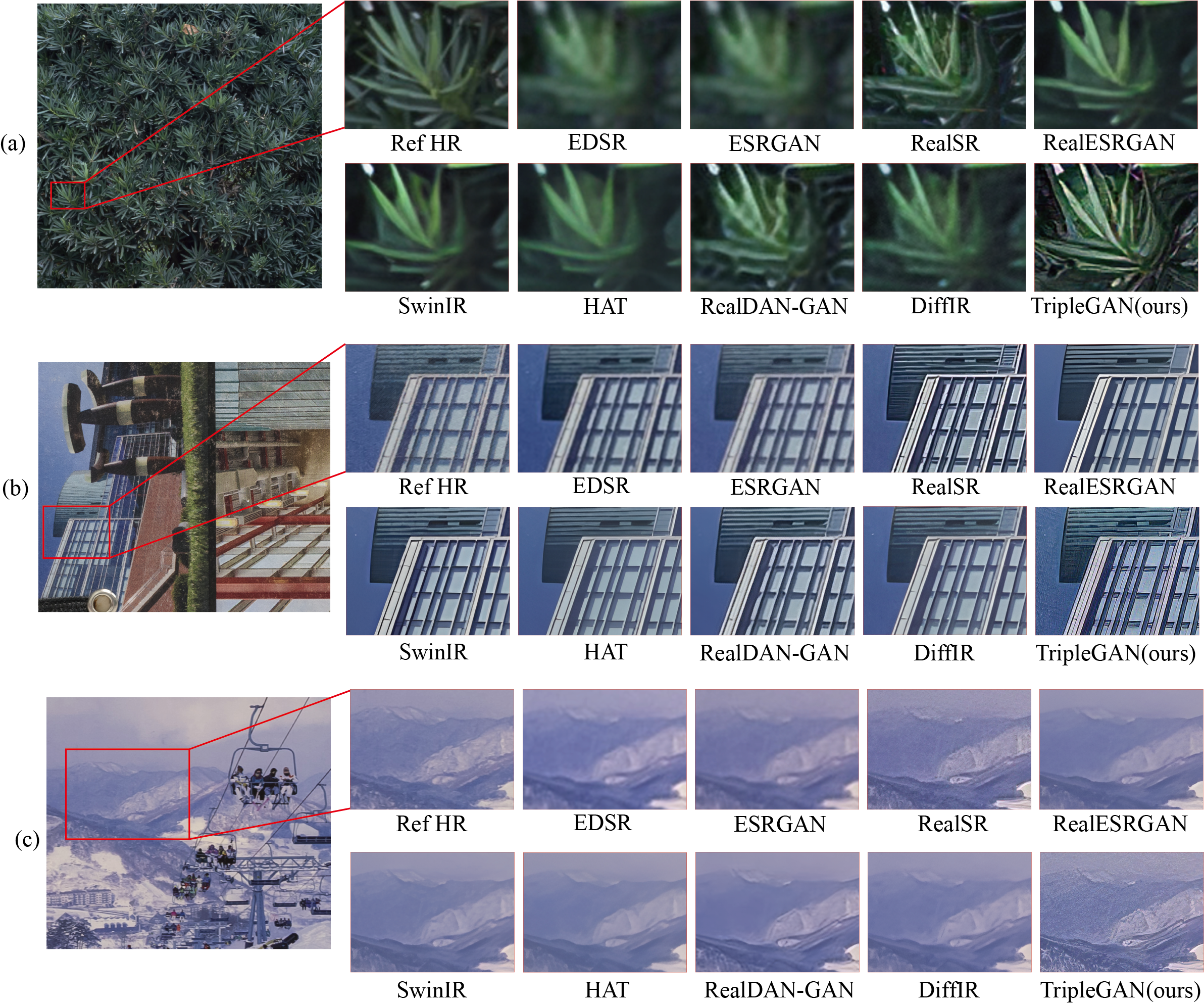}
\caption{Visual comparison on real-world LR input from RealSR dataset. Ref HR refers to the reference ground truth obtained by shooting with a different focal length.}
\label{fig_4}
\end{figure*}

For the comparative experiments, we selected the best-performing model weights of each network specifically optimized for real-world super-resolution tasks. The compared methods include Real-ESRGAN+ \cite{wang2021real}, RealSR \cite{ji2020real}, SwinIR-Real \cite{liang2021swinir}, HAT-Real \cite{chen2023activating}, RealDAN-GAN \cite{luo2023end}, DiffIR-Real \cite{xia2023diffir}, and AdaSSR \cite{fan2024towards}. The qualitative comparison results are presented in figure \ref{fig_4}, our method achieves enhanced image sharpness while avoiding excessive smoothing, striking a balance that ensures the results remain clear without losing naturalness. It preserves fine textures and intricate details better than comparative methods, effectively maintaining structural features such as leaves, buildings, and mountain patterns to deliver high-fidelity super-resolution results.

Considering the difficulty of collecting HR images ground truth in real-world scenarios, and to better evaluate super-resolution reconstruction in practical applications, we employed multiple unsupervised evaluation metrics for a comprehensive assessment. The results of the experiments comparing the mainstream models on different datasets are shown in Table~\ref{tab5}, Table~\ref{tab6}.

\begin{table}[H]
\centering
\caption{Quantitative results on the \textbf{DRealSR} ×4 test set. \\
(best metric in \textbf{bold}, second-best metric in \textit{italic}, ↓ means lower is better, ↑ means higher is better).}
\label{tab5}
\begin{tabular}{lccccc}
\toprule
Method & NIQE↓ & PIQE↓ & BRISQUE↓ & NRQM↑ & PI↓ \\
\midrule
EDSR           & 9.25 & 90.92 & 68.43 & 2.67 & 8.36 \\
ESRGAN         & 8.40 & 39.81 & 55.30 & 3.42 & 7.68 \\
Real-ESRGAN+   & 4.68 & 35.76 & 29.46 & 5.44 & 4.76 \\
RealSR-Real         & \textbf{3.66} & \textbf{16.92} & \textit{14.46} & \textbf{6.18} & \textit{3.95} \\
SwinIR-Real         & 4.67 & 40.25 & 33.39 & 5.20 & 4.88 \\
HAT-Real            & 5.33 & 35.84 & 33.92 & 5.01 & 5.34 \\
RealDAN-GAN    & 5.64 & 55.50 & 43.59 & 4.86 & 5.49 \\
AdaSSR         & 4.44 & /     & 29.24 & 5.60 & 4.78 \\
\textbf{TripleGAN (ours)} & \textit{3.69} & \textit{24.49} & \textbf{10.46} & \textit{6.16} & \textbf{3.91} \\
\bottomrule
\end{tabular}
\end{table}

\begin{table*}[!t]
\centering
\caption{Quantitative results on the \textbf{RealSR} ×4 test set. \\
(best metric in \textbf{bold}, second-best metric in \textit{italic}, ↓ means lower is better, ↑ means higher is better).}
\label{tab6}
\begin{tabular}{lccccc|ccccc}
\toprule
\multirow{2}{*}{Method} & \multicolumn{5}{c|}{\textbf{Canon x4 test}} & \multicolumn{5}{c}{\textbf{Nikon x4 test}} \\
 & NIQE↓ & PIQE↓ & BRISQUE↓ & NRQM↑ & PI↓ & NIQE↓ & PIQE↓ & BRISQUE↓ & NRQM↑ & PI↓ \\
\midrule
EDSR           & 8.64 & 90.05 & 67.68 & 2.98 & 7.92 & 8.22 & 87.90 & 65.29 & 3.10 & 7.62 \\
ESRGAN         & 8.04 & 56.71 & 57.87 & 2.92 & 7.66 & 7.42 & 61.85 & 56.37 & 3.16 & 7.23 \\
Real-ESRGAN+   & 4.38 & 39.86 & 26.09 & 6.10 & 4.21 & 4.86 & 41.54 & 31.47 & 5.70 & 4.71 \\
RealSR         & \textit{3.43} & \textbf{22.06} & \textit{19.34} & \textit{6.56} & \textit{3.51} & \textit{3.73} & \textbf{24.07} & \textit{22.36} & \textit{6.44} & \textit{3.74} \\
SwinIR-Real         & 4.45 & 45.93 & 31.79 & 5.64 & 4.53 & 4.90 & 47.16 & 35.79 & 5.35 & 4.90 \\
HAT-Real            & 5.20 & 44.22 & 32.28 & 5.35 & 5.03 & 5.32 & 40.45 & 32.18 & 5.40 & 5.12 \\
RealDAN-GAN    & 5.36 & 51.34 & 39.55 & 4.51 & 5.50 & 5.37 & 46.15 & 35.78 & 4.98 & 5.29 \\
DiffIR-Real         & 4.84 & 32.41 & 25.52 & 6.07 & 4.45 & 5.42 & 31.39 & 30.07 & 5.74 & 4.92 \\
AdaSSR         & 4.15 & /     & 23.78 & 5.69 & 3.95 & 4.34 & /     & 25.16 & 6.34 & 4.04 \\
\textbf{TripleGAN (ours)} & \textbf{3.19} & \textit{26.14} & \textbf{9.08}  & \textbf{6.92} & \textbf{3.15} & \textbf{3.63} & \textit{29.19} & \textbf{10.46} & \textbf{6.88} & \textbf{3.41} \\
\bottomrule
\end{tabular}
\end{table*}

From comparative experiments, it can be observed that under the condition of using only LR images, our TripleGAN framework achieves the best NIQE, BRISQUE, NRQM, and PI scores, and the second-best PIQE score on the RealSR test dataset ×4. In the DrealSR test dataset ×4, it achieves the best BRISQUE and PI scores, and the second-best NIQE, PIQE, and NRQM scores. These results demonstrate that, for real-world super-resolution tasks without HR ground truths, the design of degradation-consistent synthetic datasets plays a crucial role in the quality of image reconstruction. Furthermore, leveraging the proposed degradation-consistent dataset, the ThirdGAN, trained on these synthetic data and based on the ESRGAN architecture, achieves highly competitive performance, outperforming various advanced super-resolution models, including Transformer-based and Diffusion-based approaches.

Notably, our TripleGAN framework only uses a very limited number of real-world images—merely 200 LR training images per dataset. Compared with other approaches that rely on large-scale synthetic datasets generated through complex simulated degradation pipelines, our method achieves superior performance with significantly fewer data samples. This success stems from the flexibility of our model, which directly learns the degradation characteristics of the real-world LR domain and constructs customized pseudo-real domain datasets without the need for intricate or large-scale simulated degradation procedures. These datasets more accurately reflect the true degradation patterns of real-world LR images, enabling the trained super-resolution networks to generalize effectively and produce high-quality reconstruction results. Furthermore, our comparative experiments reveal that the contribution of a well-designed dataset to reconstruction performance clearly exceeds that of modifications to the super-resolution network itself. For example, although the ThirdGAN in our framework is based on the ESRGAN architecture, it achieves substantially better results than a range of representative Transformer- and Diffusion-based models when trained on our degradation-consistent synthetic dataset.

\section{Conclusion}
In this paper, we address the challenge of real-world super-resolution reconstruction under the condition where corresponding HR images for real-world LR images are difficult to obtain. We analyze the intrinsic degradation characteristics of real-world LR images and propose a TripleGAN super-resolution framework that jointly incorporates a blur degradation pattern extraction module and a domain adaptation module. 
Our findings demonstrate that the degradation domain of real-world LR images differs significantly—particularly in terms of blur characteristics—from commonly used synthetic degradation domains. A single degradation extraction network is insufficient not only for accurately modeling the blur degradation patterns of real-world LR images, but also for capturing complex and implicit degradation factors such as noise variations across different scenes and color space shifts. Likewise, relying solely on a domain adaptation network fails to bridge the blur degradation gap effectively. In contrast, the proposed TripleGAN framework jointly leverages degradation modeling and domain adaptation, significantly narrowing the blur domain gap and better aligning the overall degradation distribution with that of real-world LR images. It further enables the network to learn implicit degradation patterns, including diverse noise distributions and color domain degradations, achieving superior performance in unsupervised evaluation metrics and visually realistic reconstruction results, even under limited training data conditions.


\bibliographystyle{IEEEtran}
\bibliography{references} 

\vspace{11pt}

\begin{IEEEbiography}[{\includegraphics[width=1in,height=1.25in,clip,keepaspectratio]{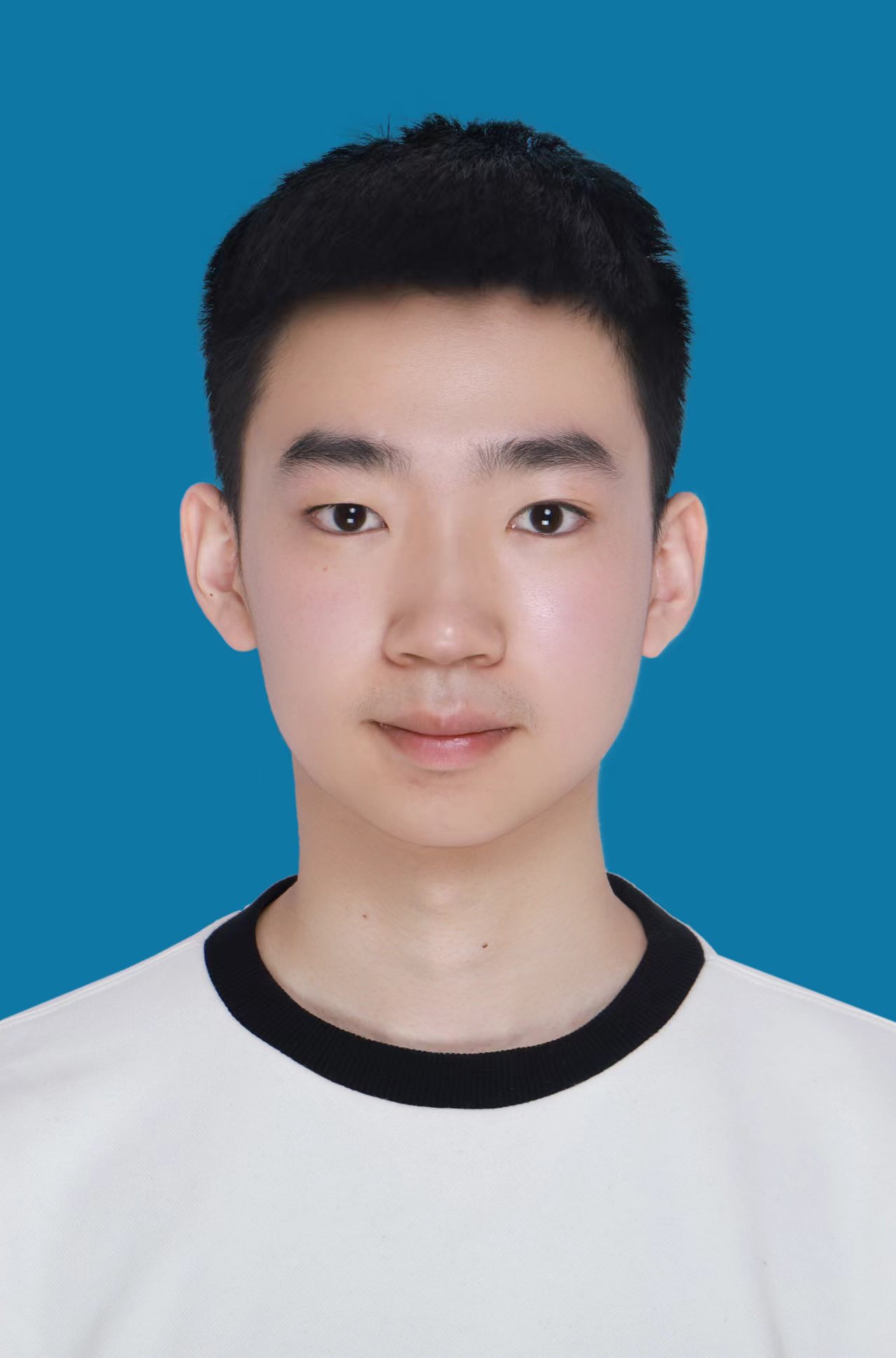}}]{Yiyang Tie}
received the B.E. degree in 2021 from Xi’an University of Technology, Xi’an, China. He is currently pursuing the M.S. degree with the School of Automation and Information Engineering, Xi’an University of Technology, Xi’an, China. His research interests include image super-resolution.
\end{IEEEbiography}
\begin{IEEEbiography}[{\includegraphics[width=1in,height=1.25in,clip,keepaspectratio]{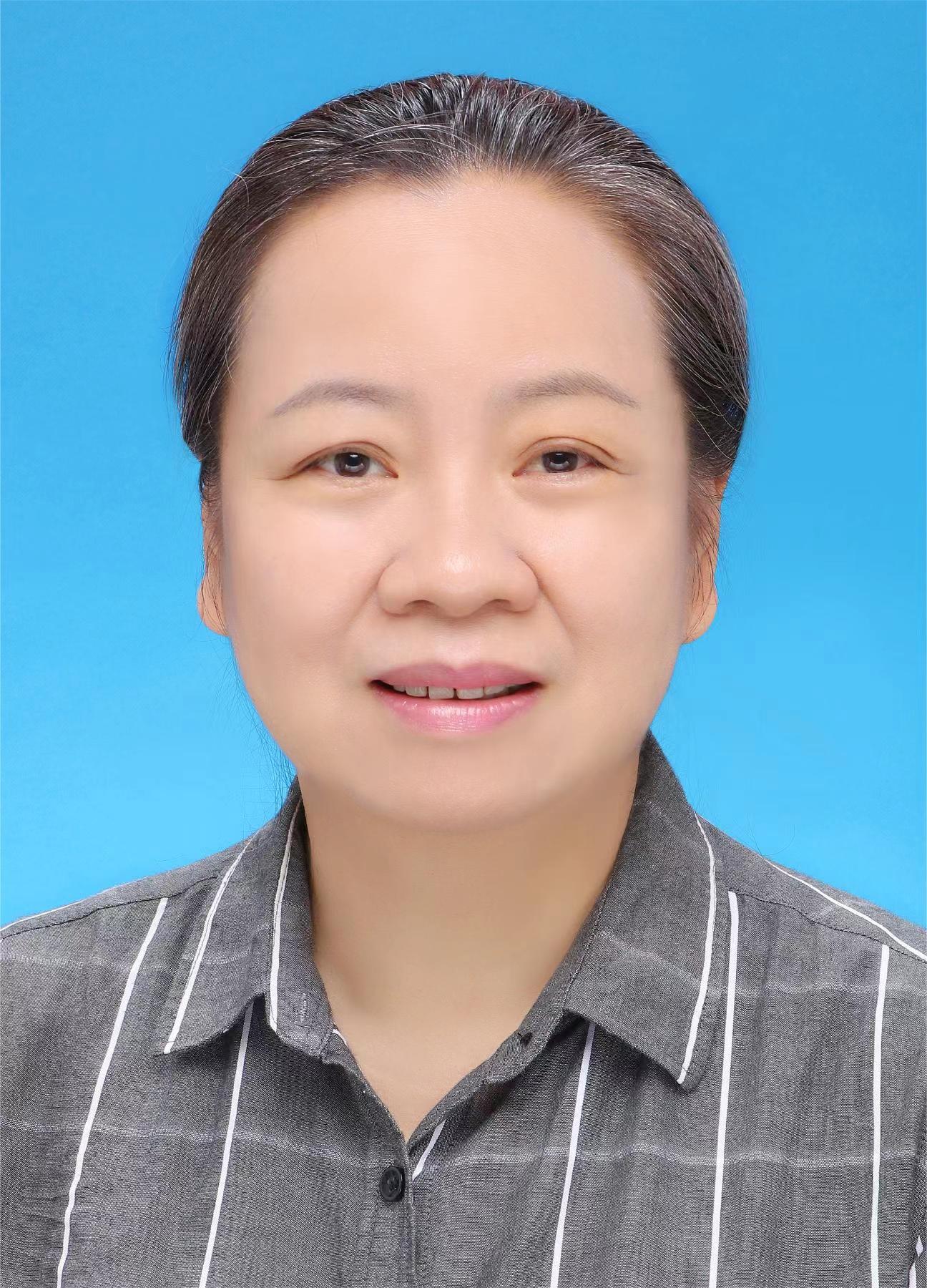}}]{Hong Zhu}
received the Ph.D. degree from Fukui University, Fukui, Japan, in 1999. She is currently a Professor with the School of Automation and Information Engineering, Xi'an University of Technology, Xi'an, China. Her research interests include image analysis, intelligent video surveillance, and pattern recognition. 
\end{IEEEbiography}
\begin{IEEEbiography}[{\includegraphics[width=1in,height=1.25in,clip,keepaspectratio]{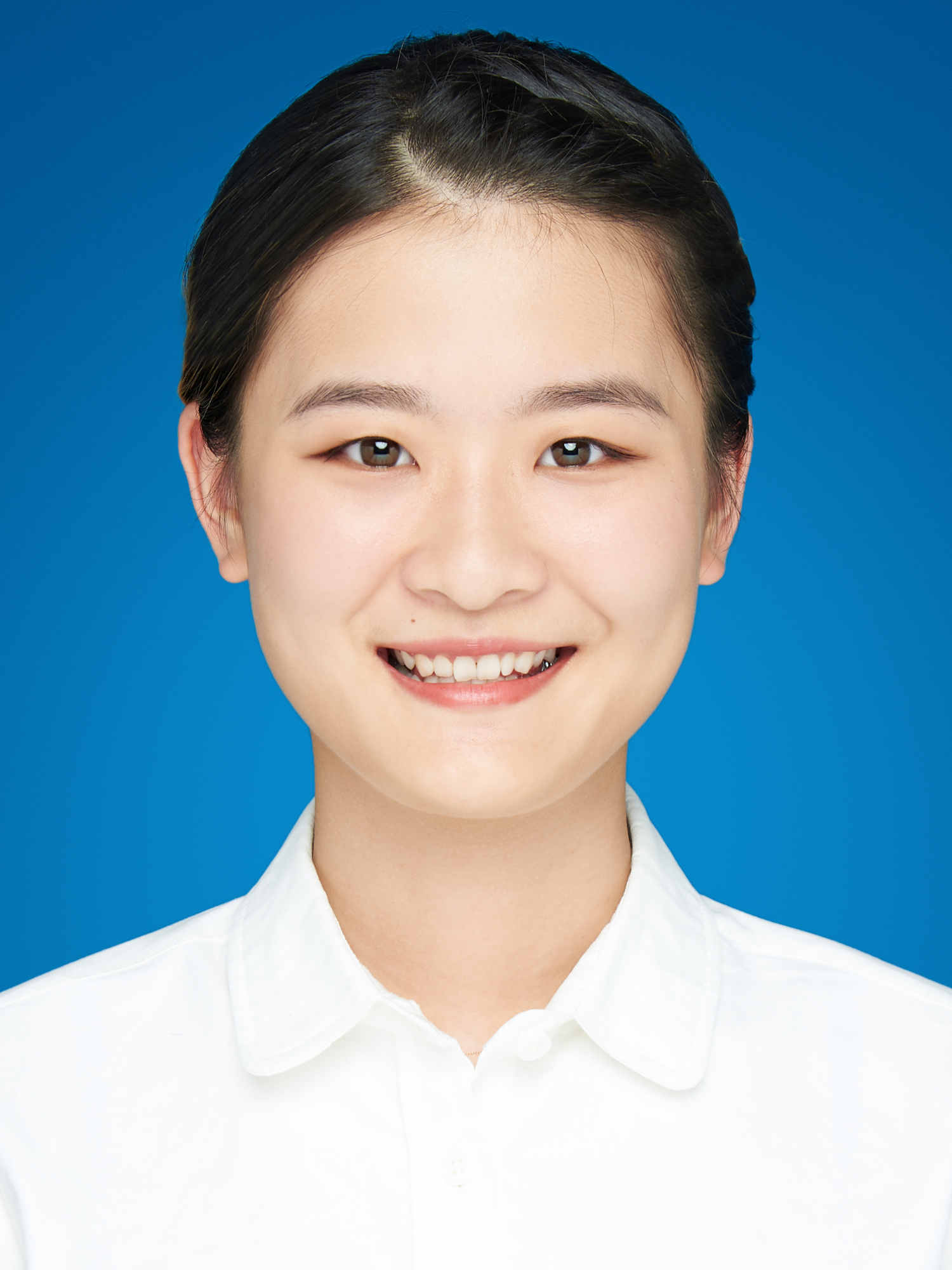}}]{Yunyun Luo}
received the B.E. degree in Engineering from Quzhou University, Quzhou, China, and the M.E. degree in Engineering from Guangxi University for Nationalities, Nanning, China. She is currently pursuing the Ph.D. degree with the School of Automation and Information Engineering, Xi’an University of Technology, Xi’an, China. Her research interests include image super-resolution.
\end{IEEEbiography}
\begin{IEEEbiography}[{\includegraphics[width=1in,height=1.25in,clip,keepaspectratio]{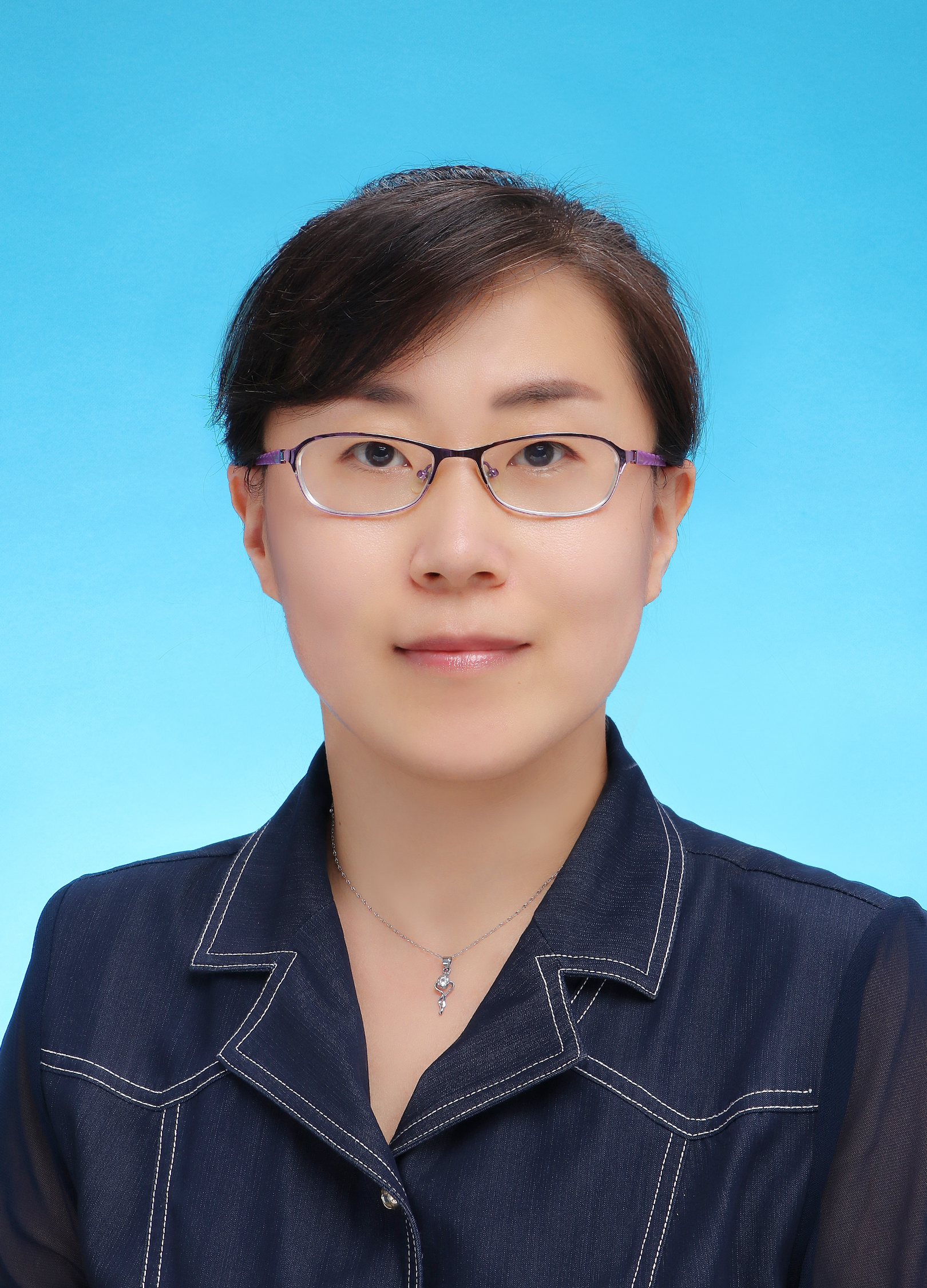}}]{Jing Shi}
received the Ph.D degree from the Institute of Automation and Information Engineering, Xi'an University of Technology, Xi'an, China, in 2019. She is currently a lecturer in the Institute of Automation and Information Engineering, Xi'an University of Technological. Her research interests include pattern recognition, scene classification and digital images processing.
\end{IEEEbiography}

\vspace{11pt}

\vfill

\end{document}